\documentclass[letterpaper, 10 pt, conference]{ieeeconf}
\IEEEoverridecommandlockouts
\usepackage{cite}
\usepackage{amsmath,amssymb,amsfonts}
\usepackage[linesnumbered, ruled]{algorithm2e}
\usepackage[flushleft]{threeparttable}
\usepackage{siunitx} 
\usepackage{graphicx}
\usepackage{textcomp}
\usepackage{algpseudocode}
\usepackage{float}
\usepackage{subfig}
\usepackage{xcolor}
\usepackage{booktabs}
\usepackage{multirow}
\usepackage{gensymb}
\usepackage{url}

\def\BibTeX{{\rm B\kern-.05em{\sc i\kern-.025em b}\kern-.08em
    T\kern-.1667em\lower.7ex\hbox{E}\kern-.125emX}}

\title{vMF-Contact: Uncertainty-aware Evidential Learning for Probabilistic Contact-grasp in Noisy Clutter
\thanks{Karlsruhe Institute of Technology, Karlsruhe, Germany. Email: {\tt\small \{name\}.\{surname\}@kit.edu}.
This work is sponsored by the DFG SFB-1574 Circular Factory project and Baden-Württemberg Ministry of Science, Research and the Arts within InnovationCampus Future Mobility. }
}

\author{Yitian Shi, 
Edgar Welte,
Maximilian Gilles,
Rania Rayyes
 }

\begin{document}
\maketitle
\thispagestyle{empty}
\pagestyle{empty}

\begin{abstract}
Grasp learning in noisy environments, such as occlusions, sensor noise, and out-of-distribution (OOD) objects, poses significant challenges. Recent learning-based approaches focus primarily on capturing aleatoric uncertainty from inherent data noise. The epistemic uncertainty, which represents the OOD recognition, is often addressed by ensembles with multiple forward paths, limiting real-time application. In this paper, we propose an uncertainty-aware approach for 6-DoF grasp detection using evidential learning to comprehensively capture both uncertainties in real-world robotic grasping. As a key contribution, we introduce vMF-Contact, a novel architecture for learning hierarchical contact grasp representations with probabilistic modeling of directional uncertainty as von Mises–Fisher (vMF) distribution. To achieve this, we analyze the theoretical formulation of the second-order objective on the posterior parametrization, providing formal guarantees for the model's ability to quantify uncertainty and improve grasp prediction performance. Moreover, we enhance feature expressiveness by applying partial point reconstructions as an auxiliary task, improving the comprehension of uncertainty quantification as well as the generalization to unseen objects. In the real-world experiments, our method demonstrates a significant improvement by 39\% in the overall clearance rate compared to the baselines. The code is available under: https://github.com/YitianShi/vMF-Contact/
\end{abstract}

%
\section{Introduction}
Recent advances in learning-based robotic manipulation have demonstrated considerable promise, primarily driven by large-scale training datasets \cite{fang2020graspnet, padalkar2023open} and the use of synthetic simulations. However, existing methods often struggle with limited adaptation and generalization in dynamic environments. For instance, in the context of a circular factory \cite{crc1574} – a manufacturing facility focused on material reuse and production design for longevity and recyclability - robots are tasked with grasping and manipulating objects under dynamic and uncertain conditions. Hence, accurate quantification of uncertainty is crucial for reliable performance where robotic operations face challenges such as occlusions, sensor noise, view perspective, and the presence of OOD objects. To address these, effective modeling and disentanglement between aleatoric uncertainty, which stems from inherent sensor noise, and epistemic uncertainty that arises from the distribution shift, is essential for improving grasp success rates and ensuring safety in real-world applications. 
\begin{figure}
        \includegraphics[width=.98\linewidth]{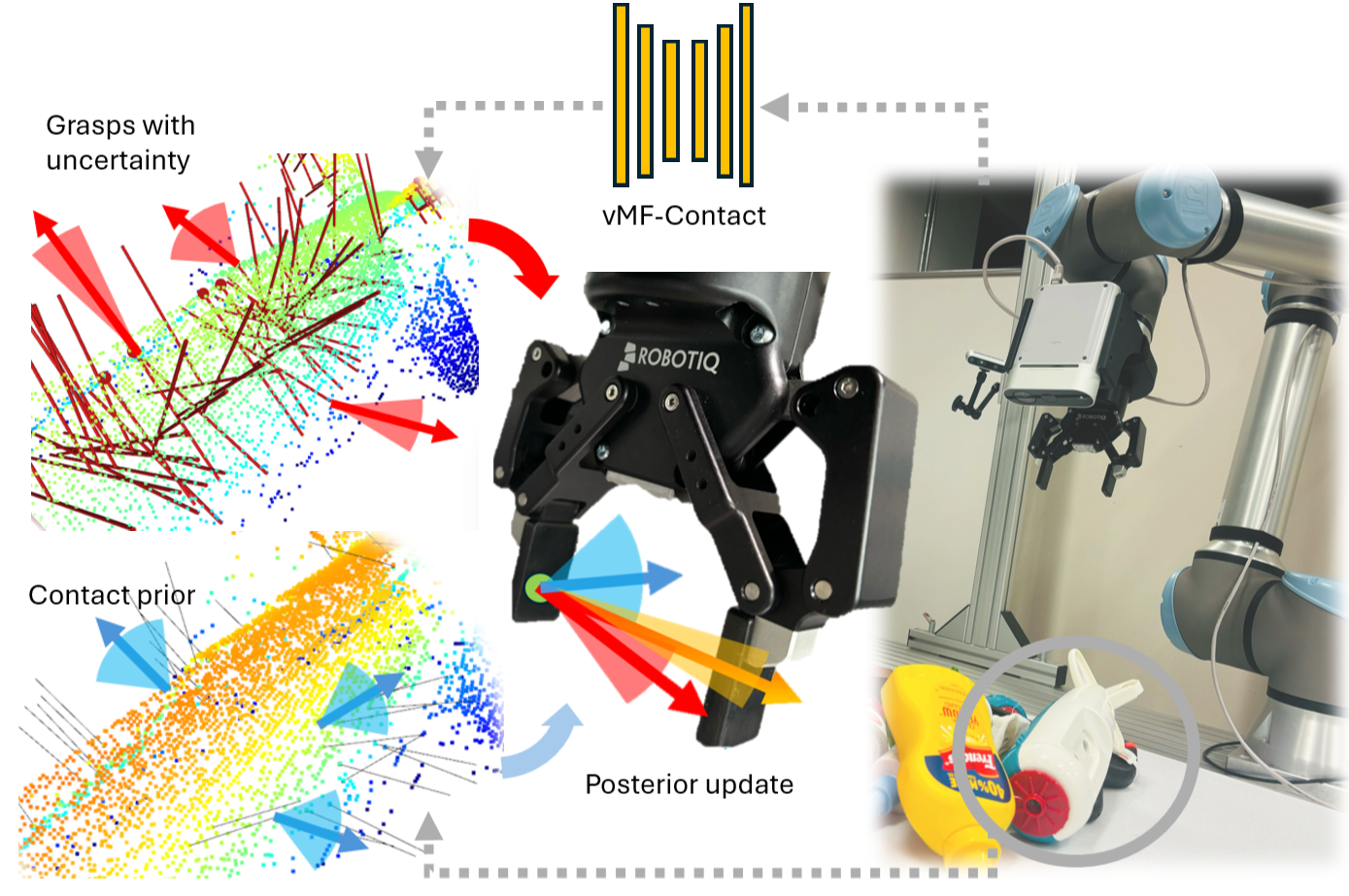}
            \caption{Inference pipeline for vMF-Contact in real-world based on posterior update.}
            \label{fig:f}
\end{figure} 

For accurate uncertainty quantification, evidential deep learning \cite{sensoy2018evidential} introduces second-order optimization techniques to achieve precise, deterministic uncertainty estimation. In contrast to Bayesian Neural Networks (BNNs), where sampling-based or variational inferences introduce computational overhead\cite{gawlikowski2023survey}, evidential deep learning provides an efficient alternative by comprehensive uncertainty estimation in a single forward path. Specifically, Dirichlet-based evidential models have demonstrated considerable potential for OOD detection in classification \cite{sensoy2018evidential, ancha2024deep}. Furthermore, natural posterior networks (NatPN) 
\cite{charpentier2020posterior} extend the evidential learning by modeling arbitrary posteriors within the exponential family distributions.
Based on this line of approaches, we develop uncertainty-aware grasp detectors, ensuring reliable performance in real-world environments. 

In summary, as the main contributions: (1) We develop the vMF-Contact, a novel architecture that facilitates probabilistic contact-based grasp learning with posterior updates. (2) We perform theoretical analysis for vMF Contact through an analytically formulated second-order objective, which integrates evidential learning for precise directional uncertainty estimation, ensuring reliable grasp predictions with carefully designed informative prior. 
(3) To improve the geometric comprehension of the encoder, we employ auxiliary point reconstruction task that enhances both uncertainty quantification and grasp success. In our experiments, we systematically selected and compared existing PointNet-based architectures to evaluate the generalizability of our framework. Our method demonstrate reliable performance in real-world experiments, showing substantial improvements in grasp performance and robustness against OOD scenarios without any sim-to-real adaptation.

\section{Related works}

\subsection{Deep learning for SE(3) robotic grasping}
Learning-based SE(3) robotic grasping approaches have shown significant achievement in the real world \cite{newbury2023deep}. For feature extraction in supervised grasp learning, one class of approaches leverages 3D convolution over volumetric input representations such as Truncated Signed Distance Functions (TSDF) \cite{breyer2021volumetric, jauhri2023learning} or rendering-based approaches \cite{dai2023graspnerf, jauhri2023learning}, requiring multi-view input to achieve comprehensive scene understanding while balancing the trade-off between resolution and computational cost. In contrast, PointNet-based architectures \cite{sundermeyer2021contact, ni2020pointnet++} operate directly on point clouds without the need for voxelization, alleviating this overhead while retaining geometric fidelity, offering a more efficient alternative for grasp detection tasks. However, in highly cluttered and dynamic environments, naive end-to-end methods are susceptible to OOD observations, leading to notable performance degradation. Therefore, capturing and quantifying uncertainty is essential to mitigate these impacts and ensure a more robust grasp inference \cite{shi2024uncertainty}.
\subsection{Evidential learning for  uncertainty estimation}
Deep uncertainty estimation has been extensively studied for over a decade \cite{gawlikowski2023survey, kendall2017uncertainties}. Early efforts focused on variational Bayesian inference by back-propagation \cite{blundell2015weight} or dropout \cite{gal2016dropout} have faced challenges in faithful representation of epistemic uncertainty \cite{izmailov2021bayesian, seligmann2024beyond}. In contrast, ensemble approaches \cite{lakshminarayanan2017simple, seligmann2024beyond} have demonstrated significant improvement in enhancing both the prediction accuracy and the reliability of uncertainty estimate 
\cite{Osband2016DeepDQN} with the cost of multiple forward paths and limited model samples, requiring moderate pruning and distillation. 

Due to these trade-off and robustness against OOD recognition, recent studies in evidential learning \cite{sensoy2018evidential} opt to second-order risk minimization on the distributional parameterizations. Specifically, given data \(D\) and observation \(x\), the Bayesian inference in evidential learning follows: \[
\text{P}(y \mid x, D) = \int \underbrace{\text{P}(y \mid \mathbf{\theta})}_{\text{aleatoric}} \underbrace{\text{P}( \mathbf{\theta} \mid x, D)}_{\text{epistemic}} \, d\mathbf{\theta}
\]
here epistemic uncertainty is captured through the variability of the posterior distribution \( \text{P}( \mathbf{\theta} | x, D) \),  which indicates the model’s confidence in its predictions for different values of \(  \mathbf{\theta} \) \cite{ulmer2021prior}. 
 Aleatoric uncertainty is quantified by the entropy under the likelihood distribution \( \text{P}(y |  \mathbf{\theta}) \), reflecting uncertainty inherent in the data noise.  

Among these, Charpentier et al. \cite{charpentier2020posterior} introduced natural posterior networks (NatPN), a general framework for predictive posteriors that is tailored to general exponential distributions. They leverage the normalizing flows \cite{papamakarios2021normalizing} that form a bijective mapping between the encoder features and a single-dimensional isotropic Gaussian, facilitating the density quantification and OOD recognition with a single forward path,  regardless of any learning discrepancy caused by additional data. Moreover, the practical implementation of NatPN in object detection \cite{feng2023topology} and semantic segmentation \cite{ancha2024deep} highlights the following key effects that may improve its uncertainty estimation: i) expressiveness and complexity of the encoder features through e.g. auxiliary reconstruction tasks and ii) pre-trained base distribution for normalizing flows to alleviate the topology mismatch. Therefore, we see great potential of evidential learning for robotic grasping.

\subsection{von Mises-Fisher distribution}
In the context of robotic manipulation, directional data play a critical role in modeling object orientations and robot movements. As one of the exponential family distributions, the von Mises-Fisher (vMF) distribution is widely applied in statistics and deep learning, such as surface normal estimation \cite{bae2021estimating}, face recognition \cite{hasnat2017mises}, and the distillation of foundation model features \cite{govindarajan2024dino}. In robotic grasping, \cite{liu2024efficient} employed the Power Spherical (PS) distribution \cite{de2020power} to model baseline vectors in contact grasp \cite{sundermeyer2021contact}, favored over the vMF due to its stability in sampling. However, we identify that the posterior update for the PS distribution can only be expressed in closed form, introducing limitations in terms of analytical tractability for Bayesian inference.

\section{Evidential learning for contact grasp distribution}
To effectively address OOD challenges in robotic grasping, modeling epistemic uncertainty is crucial. Hence, we explore evidential learning within a probabilistic contact grasp, enabling us to model both aleatoric and epistemic uncertainties.

\subsection{Problem formulation}

Given a raw point cloud \(N^3\) from depth sensors, we aim to generate a comprehensive set of potential contact grasps, formalized as tuples \(\{g_n = (\mathbf{c}, \mathbf{b}, \mathbf{a}, w)_n | n\in N\}\) that are conditioned on the sampled points \(n\) (cf.~Fig.~\ref{fig:pm}). 1) \(\mathbf{c}\in \mathbf{R}^3\) denotes the queried potential contact points. 2) The baseline vector \(\mathbf{b} \in \mathbf{R}^3\) captures the spatial relation between the observable contact point \(\mathbf{c}\) and the inferred (latent) contact point corresponding to the opposing finger of the gripper. 3) The approach vector \(\mathbf{a}\in \mathbf{R}^3\) describes the gripper's motion direction to approach the object. 4) The grasp width
\(w\in R\) defines the distance between the two gripper fingers, determined by the spatial distribution of the contact points. One significant advantage of contact grasp representation is its hierarchical nature, which aligns with Bayesian modeling. Given contact point \(c\), we follow the hierarchical decomposition by \cite{liu2024efficient}, breaking down the hierarchical contact grasp into each factor:
\begin{align*}
\text{P}(g | \mathbf{c}) = \text{P}(w, \mathbf{a}, \mathbf{b} | \mathbf{c}) \propto \text{P}(w | \mathbf{b}, \mathbf{c}) \mathbf{P}(\mathbf{a} | \mathbf{b}, \mathbf{c}) \text{P}(\mathbf{b} | \mathbf{c})
\end{align*}
assuming independence between \(w\) and approach vector \(\mathbf{a}\). %

Nevertheless, \cite{liu2024efficient} focuses on modeling the conditional baseline distribution (likelihood) using a Power-Spherical (PS) density: \(\text{P}(\mathbf{b} | \mathbf{c}) = \mathbf{PS} (\mathbf{b} | \mu_\mathbf{c}, \kappa_\mathbf{c})\) which only consider aleotoric uncertainty. In contrast,  we model both aleatoric and epistemic uncertainty through a principled hierarchical posterior update. 
We argue that the conjugate prior for the PS distribution is analytically intractable, making it challenging to rigorously capture epistemic uncertainty in a principled manner. Hence, we adopt the vanilla vMF likelihood to model the baseline vector and provide the analytical formulation of the posterior distribution in the following section.

\subsection{vMF distribution as natural posterior}
\begin{figure}
\vspace{2mm}
\centering
        \includegraphics[width=.95\linewidth]{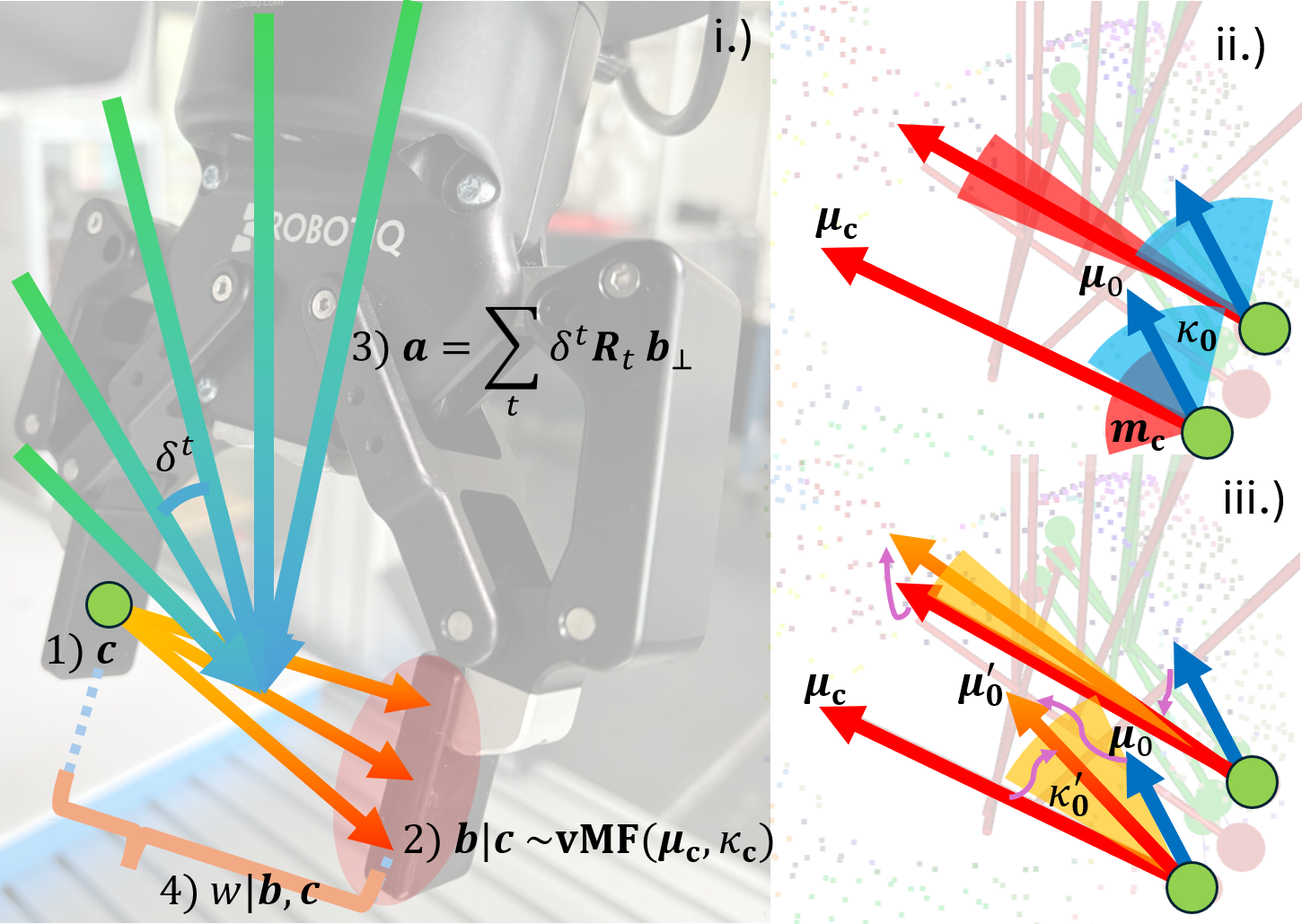}
            \caption{Illustration of posterior update for grasp baseline direction wrt. predictive distributional uncertainty. We denote \(R_{t}\) as the rotation matrix to transform \(\mathbf{b}\) to the \(\mathbf{t}\)-th bin.}
            \label{fig:pm}
\end{figure} 
The baseline vector \(\mathbf{b}\) is modeled as 3D vMF likelihood:
\begin{align*}
p(\mathbf{b}|\mathbf{c}) = \mathrm{vMF}(\mathbf{b}|  \mathbf{\mu_\mathbf{c}}, \kappa_\mathbf{c}) &=  Z(\kappa_\mathbf{c}) \exp(\kappa_\mathbf{c}  \mathbf{\mu_\mathbf{c}}^\top \mathbf{b}),
\end{align*}
with \(Z(\kappa_\mathbf{c}) = \kappa_\mathbf{c}^{-1}{4\pi \sinh(\kappa_\mathbf{c})}\) for normalization.  \(\mathbf{\mu_\mathbf{c}}\) is the mean direction, \(\kappa_\mathbf{c}\) is the concentration of the vector density \cite{mardia1976bayesian}. This can be interpreted as a Gaussian for unit vectors on the 3D sphere given initial contact point \(\mathbf{c}\). For each \(\mathbf{c}\), the corresponding parametrizations can be inferred by a deep decoder (e.g. a multi-layer perceptron): \( (\mathbf{\mu_\mathbf{c}}, \kappa_\mathbf{c}) = \mathrm{MLP}(z_\mathbf{c})\) on the pointwise feature \(z_\mathbf{c}\) (cf.~Fig.~\ref{fig:c3d}, Sec:~\ref{sec:vMF}).

The conjugate prior of vMF likelihood is modeled as another vMF: \(\mathbf{\mu_\mathbf{c}} \sim \mathrm{vMF}(\mathbf{\mu_\mathbf{c}} | \mu_0, \kappa_0)\), given deterministic \(\kappa_\mathbf{c}\).
Since vMF distribution belongs to the exponential family, the parameters for posterior distribution \( \mathrm{vMF}(\mu_\mathbf{c} | \mu'_0, \kappa'_0)\) can be updated as following \cite{charpentier2021natural}:
\begin{align*}
    \mu'_0 = \frac{\kappa_0 \mu_0 + m_\mathbf{c}  \mu_\mathbf{c}}{\kappa_0 + m_\mathbf{c}}, \quad \kappa'_0 = \kappa_0 + m_\mathbf{c}
\end{align*}
 \(m_\mathbf{c}\) is the evidence ("pseudo-count"), representing the belief on observed likelihood parameter \(\mu_\mathbf{c}\) as a discrepancy measure compared to training data, reflecting the epistemic uncertainty. As an intuitive explanation of the posterior update, Fig.~\ref{fig:pm} (right) illustrates the impact of evidence \(m_\mathbf{c}\) on the final posterior \(\mu'_0\), which can be understood as an interpolation between the prior \(\mu_0\) and the observed mean direction \(\mu_\mathbf{c}\) weighted by \(\kappa_0\) and \( m_\mathbf{c}\) respectively. For small \(m_\mathbf{c}\) the posterior tends to stay conservative aligning with the prior \(\mu_0\). Conversely, by increasing evidence belief \( m_\mathbf{c}\), the posterior is more confident in the predictive likelihood \(\mu_\mathbf{c}\). 

Notably, for joint conjugate prior \( \textbf{}p(\mu_\mathbf{c}, \kappa_\mathbf{c} )\), there's no analytical solution for exact posterior density \cite{nunez2005bayesian} to the best of our knowledge. Hence, we stay with the deterministic likelihood concentration \(\kappa_\mathbf{c}\) that models the aleatoric uncertainty.

\subsection{Bayesian loss for vMF posterior}
Here we aim to analyze the feasibility of analytical second-order objective for optimizing the vMF posterior parametrizations. Given ground truths \(\tilde{\mathbf{b}}_\mathbf{c}\) and predictive likelihood  \( (\mathbf{\mu_\mathbf{c}}, \kappa_\mathbf{c}) = \mathrm{MLP}(z_\mathbf{c})\) decoded from point feature \(z_\mathbf{c}\), the optimization for general exponential posterior can be formulated as "Bayesian loss" following \cite{charpentier2020posterior}:
\begin{align}
\mathcal{L}^{\text{Bl}}_{\mathbf{b}_\mathbf{c}} = & -  \underbrace{\mathbf{E}_{ \mathbf{\mu}_\mathbf{c} \sim \mathbf{Q}_{\text{post.}  }} \left[ \log \text{p}(\tilde{\mathbf{b}}_\mathbf{c} \mid  \mathbf{\mu}_\mathbf{c}, \mathbf{\kappa}_\mathbf{c}) \right]}_{\text{(i)}} - \gamma * \underbrace{\mathbb{H}[\mathbf{Q}_{\text{post.}}]}_{\text{(ii)}}
\label{Bl}
\end{align} 
where (i) is the expected log-likelihood and (ii) denotes the entropy of the posterior predictive distribution \(\mathbf{Q}_{\text{post.}} = \mathrm{vMF}(\mu_\mathbf{c} | \mu'_0, \kappa'_0) \) discounted by \(\gamma \). For vMF distributions, the calculation of term (i) gives:
\begin{align*}
 \int_{\mu_\mathbf{c} \in S^2} Z(\kappa'_0) \exp(\kappa'_0 \mu_0'^{\top} \mu_\mathbf{c}) \left( \log Z(\kappa_\mathbf{c}) + \kappa_\mathbf{c}\tilde{\mathbf{b}}_\mathbf{c}^{\top} \mu_\mathbf{c} \right) d\mu_\mathbf{c}
\end{align*} 
We may assume \( \mu'_0 = (0, 0, 1)^\top \) to simplify the calculation given the rotational symmetry of integration on \(\mu_c\) over the unit sphere \(S^2\). In this case, the relative orientation between \(\tilde{\mathbf{b}}_\mathbf{c}\) and \(\mu'_0\) still need to be preserved. 
We provide the analytical formulation\footnote{The proof is omitted due to space constraints} as:
\begin{align}
    \text{(i)} = \log Z(\kappa_\mathbf{c}) + \left( \tanh^{-1}( \kappa'_0) - {\kappa'_0}^{-1} \right) \kappa_\mathbf{c}  {\tilde{\mathbf{b}}_\mathbf{c}}^\top{ \mu'_0}
    \label{ell}
\end{align}
For the minimization of Eq. \ref{Bl}, this term intends to reduce the epistemic uncertainty (increase \( {\kappa'_0}\)) by maximizing (\( \tanh^{-1}( \kappa'_0) - {\kappa'_0}^{-1}\)). While this is balanced through the maximization of the term (ii), as the posterior's entropy \(\mathbb{H} (\mathrm{vMF}( \mu_\mathbf{c} | \mu'_0, \kappa'_0))\). This can be derived as \cite{mardia1976bayesian}:
\begin{align}\label{entropy}
    \text{(ii)}= -\log  Z(\kappa'_0) - \kappa'_0 * \tanh^{-1}(\kappa'_0) + 1.
\end{align}

Notably, in Bayesian modeling, an isotropic zero-mean infinite-variance Gaussian \cite{carlin2008bayesian} is often used as a non-informative prior. For the prior \(\mathrm{vMF}( \mu_\mathbf{c} | \mu_0, \kappa_0)\), a common assumption is to set a small concentration parameter \(\kappa_0 \approx 0\) with an arbitrary mean direction \( \mu_0\), reflecting a weakly informative prior. However, in the context of grasp learning, where force closure constraints are critical, we opt for an informative prior by aligning the mean direction \( \mu_0\) with the negative surface normal of point \(\mathbf{c}\) on the object as \(\mu_0(\mathbf{c}) = -\mathbf{n}_{\mathbf{c}}\), reflecting the expected support of contact forces onto the object surface that fulfills antipodality constraints. We take \(\kappa_0 = 1\) as constant prior concentration.

\subsection{vMF-Contact for evidential learning}
\label{sec:vMF}
\begin{figure*}[htbp]%
\vspace{2mm}
    \centering
        \includegraphics[width=.92\linewidth]{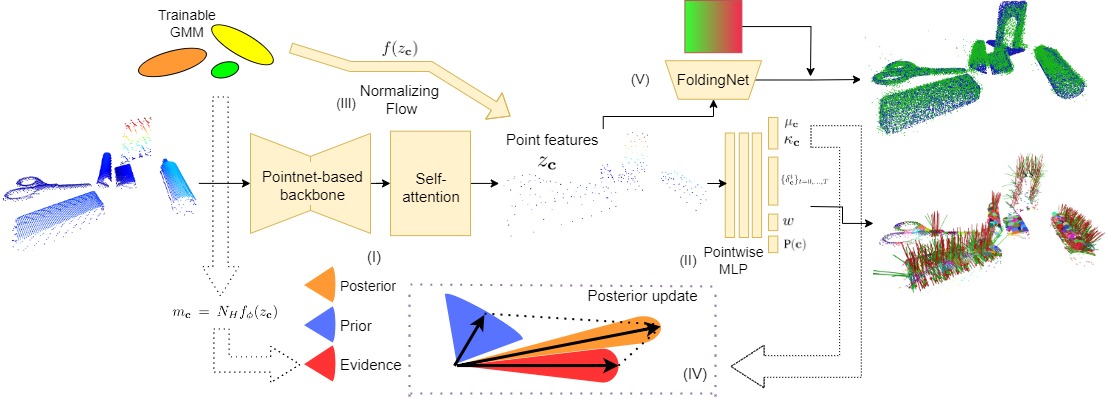}
    \caption{vMF-Contact architecture. The raw point clouds from a single camera view go through the PointNet-based backbone (I), where the feature is further enhanced by geometric-aware self-attention. The down-scaled point features are taken by (II) pointwise linear layers to predict the conditional grasp orientations. A residual flow is trained to estimate the density of point features (III) and provide the evidence that serves the posterior update (IV). Auxiliary point completion (IV) is applied using a shared folding net to enhance the feature expressiveness (green points) and robustness against input noises supervised by "ground truths" point clouds (blue).}
\label{fig:c3d}
\end{figure*} 
vMF-Contact is built upon NatPN and able to enhance feature learning for probabilistic contact grasp through a Bayesian treatment. To ensure seamless integration with point cloud-based learning, our vMF-Contact architecture is designed to incorporate evidential learning to capture the uncertainty under contact grasp representation.

Fig.~\ref{fig:c3d} illustrates vMF-Contact architecture as follows: 
(I) Given \(N=20000\) points from a single camera view, a PointNet-based backbone with concatenated self-attention layers downscales the scene into \(N'=2000\) point features \(z_\mathbf{c}\), as similar to Contact-GraspNet \cite{sundermeyer2021contact}. 
(II) A shared pointwise linear layer outputs the grasp parametrizations including conditional baseline likelihood \( \mathbf{\mu_\mathbf{c}}, \kappa_\mathbf{c}\), quantized bin scores \(\{\delta^t_\mathbf{c}\}_{t=0,...,T}\) for calculating approach vector \(\mathbf{a}_\mathbf{c}\), grasp width \(w_\mathbf{c}\) and the grasp quality \(\text{P}(\mathbf{c})\). Specifically, bin scores $\delta^t_\mathbf{c}$ are predicted for each quantized bin over all the possible rotation angles by seperating \(0-180\degree\) into \(T\) bins, which are perpendicular to the baseline vectors  (cf.~Fig.\ref{fig:pm} i.). Then a normalizing flow \(f_{\phi}\) attempts to fit a trainable GMM to the point features (III) following \cite{charpentier2021natural}: \(m_\mathbf{c} = N_H f_{\phi}(z_\mathbf{c})\), where the estimated feature density \(f_{\phi}(z_\mathbf{c})\) is mapped to the exact evidence \(m_\mathbf{c}\) scaled by the factor \(N_H\), yielding the epistemic uncertainty. Finally, the predictive evidence serves the posterior update (IV) on our informative prior.

Moreover, recent works \cite{ancha2024deep, charpentier2023training} have emphasized the importance of enhancing feature understanding in posterior networks through two key strategies: First, the complexity of the encoder architecture and the expressiveness of its features generally lead to improved uncertainty quantification. Based on this, we selected PointNeXt as our backbone following a comprehensive comparison with state-of-the-art PointNet-based encoders 
in our experiments (Sec. \ref{sec:exp}). 
Second, to adapt an analogous auxiliary reconstruction task \cite{ancha2024deep} to the 3D PointNet-based architecture, we employed a shared FoldingNet \cite{yang2018foldingnet} to reconstruct the local point clouds around the generated contact points (V), using the feature with feature \(z_\mathbf{c}\). This aims to extend the pixel-wise reconstruction from 2D features as suggested by \cite{ancha2024deep} to the point-based domain. We supervise the reconstruction results by uniformly sampled points from the object model without occlusion, which aims to improve robustness against input noise while emphasizing local geometric details. It is worth noting that, 
our reconstruction head is used solely as auxiliary supervision, which is not involved during inference.

\subsection{Optimization}
\label{opt}
Given ground truths \(\{\tilde{g}_n = (\tilde{\mathbf{c}}, \tilde{\mathbf{b}}, \tilde{\mathbf{a}}, \tilde{w})_n \mid n \in N\}\), we aim to optimize each element of the grasp representation. Due to the imbalance between the number of ground truth grasps and the predictions, each prediction is matched with its nearest neighbor within 2cm. The grasp score on point \(\mathbf{c}\) is denoted as \(\text{P}(\mathbf{c})\) to reflect the grasp quality accounting for collisions and force closure. The unpaired predictions will considered as negative samples (\(\text{P}(\mathbf{c}) = 0)\)). In addition to the conventional L1-regression for grasp width \(\mathcal{L}_{w_\mathbf{c}} = \|\tilde{w}_\mathbf{c} - w_\mathbf{c}\|\) and binary cross-entropy for grasp success \(\mathcal{L}_{\mathbf{c}} = \text{BCE}(\tilde{\mathbf{c}}, \mathbf{c})\), the following key changes from existing works:

\subsubsection{Baseline Vector}
To compare with deterministic and first-order objectives, the baseline vector loss \(\mathcal{L}_{\mathbf{b}_\mathbf{c}}\) for each sample \(g_\mathbf{c}\) is optimized using one of the following methods:
\begin{itemize}
    \item \textbf{Cosine slimilarity loss}:
    \(
    \mathcal{L}^{\text{cosine}}_{\tilde{\mathbf{b}}_\mathbf{c}} = 1 -\tilde{\mathbf{b}}_\mathbf{c}^\top  \mu_\mathbf{c}
    \)

    \item \textbf{Negative log-likelihood}:
    \(
    \mathcal{L}^{\text{nll}}_{\tilde{\mathbf{b}}_\mathbf{c}}=-\log \mathrm{vMF}(\tilde{\mathbf{b}}_\mathbf{c}| \mu_\mathbf{c}, \kappa_\mathbf{c})
    \)

    \item \textbf{Bayesian loss}:
    \(\mathcal{L}^{\text{Bl}}_{\tilde{\mathbf{b}}_\mathbf{c}} = \text{Eq. (\ref{ell}) + Eq. (\ref{entropy})}
    \)
\end{itemize}

\subsubsection{Approach Vector}
For the approach vector, unlike existing work\cite{liu2024efficient} which uses binary bin scores \(\delta^t _\mathbf{c}\in \{0, 1\}\) to represent grasp collision, we utilize the cosine similarity between the ground truth approach vector \(\tilde{\mathbf{a}}_\mathbf{c}\) and each bin vector 
to supervise soft bin score prediction, formulated as: \(
\mathcal{L}_{\mathbf{a}_\mathbf{c}} = \sum_{t=0}^{T} ||\delta^t _\mathbf{c}- \tilde{\mathbf{a}}_\mathbf{c}^\top \mathbf{a}_{\delta^t}||^2
\). This allows for soft assignment of predicted approach bin scores, transforming the task from binary classification to a regression problem.

\subsubsection{Flow and reconstruction}
To learn a normalizing flow, the forward Kullback-Leibler divergence (KLD) is applied between the predicted density and a trainable Gaussian Mixture Model (GMM) with 20 components for density estimation \(
\mathcal{L}_{\phi} = \text{KLD}(f_\phi (z_\mathbf{c}), \text{GMM})\). Residual flows \cite{chen2019residual} supported by existing library \cite{Stimper2023} is used that realizes the trainable base GMM distribution. The reconstruction loss  \(\mathcal{L}^{rec}_\mathbf{c} \) is trained by the extended Chamfer Distance \cite{yu2021pointr}, \(P_\mathbf{c}\) refers to the reconstructed points and \(Q_\mathbf{c}\) the local patch of ground-truth points around point \(\mathbf{c}\):
\[
\mathcal{L}_\mathbf{c}^{\text{rec}}(P_\mathbf{c}, Q_\mathbf{c}) = \sum_{p \in P_\mathbf{c}} \min_{q \in  Q_\mathbf{c}} \| p - q \|_2^2 + \sum_{q \in  Q_\mathbf{c}} \min_{p \in P_\mathbf{c}} \| q - p \|_2^2
\]

\begin{figure}[tb!]
\vspace{2mm}
    \centering
        \includegraphics[width=.98\linewidth]{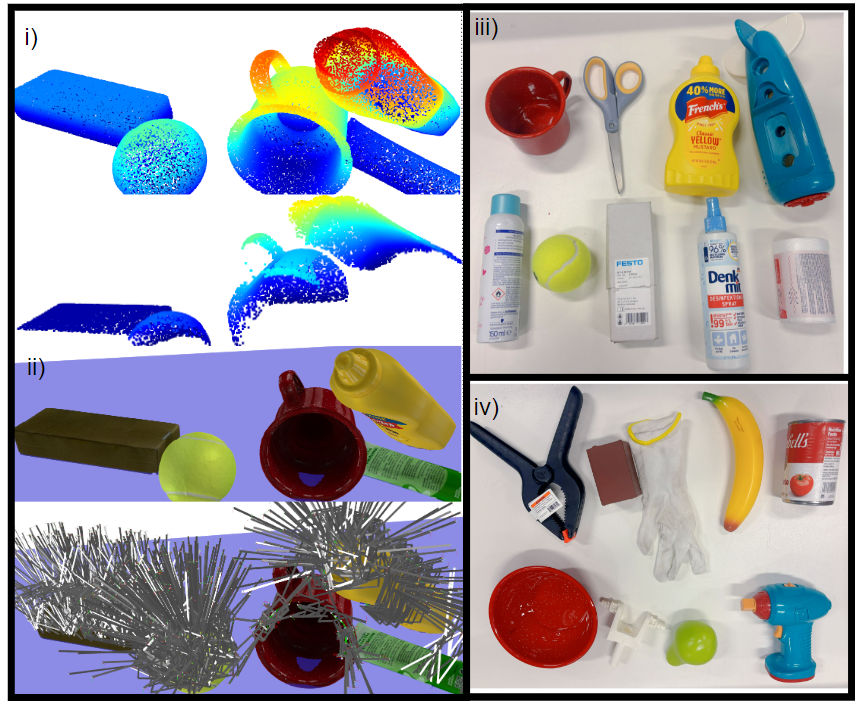}
    \caption{ i) Generated data from simulation with ground truths point clouds ii) Ground truths grasps in the scene iii) ID objects iv) OOD objects}
            \label{fig:data}
\end{figure} 
\section{Experiments}\label{sec:exp}
We attempt to answer the following questions: i) How do the complexity of the backbone architectures and choice of optimization objectives affect the ability of vMF-Contact to capture uncertainty?
ii) Does the point-based reconstruction task enhance feature expressiveness, particularly by improving the accuracy of uncertainty estimation and increasing grasp success rates?
iii) What is the model's overall generalization ability with regard to OOD scenes and objects?

\subsection{Data generation}\label{sec:data_gen}
In line with the recent trend of simulation-based data generation, we leverage Isaac Sim to generate realistic scenes, enabling efficient parallelization. We select small object set to show the generalization of our method to novel scenarios. Our object set consists of only 12 objects, which are selected from MetaGraspNet (MGN)\cite{10309974}, including items from YCB benchmark \cite{calli2015ycb} and logistics domains for increased diversity. To evaluate OOD generalization of vMF-Contact, we generated 1600 distinct scenes by randomly select a subset from our object set (with a 0.5 probability). 
In each parallelized environment in simulation, the objects were instantiated with randomized poses and appearance, ensuring a wide variety of arrangements. For each scene, we recorded the following data as one package (cf. Fig~\ref{fig:data} i), ii)): 1) noisy point clouds from randomized camera viewpoints, 2) ground truths point clouds, which refer to the uniform point samples over the object surface that serves the auxiliary point reconstruction and 3) The gound truths grasps sampled and analyzed by the collision checker over the scene using simulated Franka parallel jaw gripper for simplicity.
\subsection{Training details}
Our loss objective is a weighted sum of the individual components in Sec. \ref{opt}: \(
\mathcal{L}_{\text{total}} = \frac{1}{N} \sum_{i=0}^N (\lambda_w \mathcal{L}_{w_\mathbf{c}} + \lambda_\mathbf{c} \mathcal{L}_\mathbf{c} + \lambda_b \mathcal{L}_{\mathbf{b}_\mathbf{c}} + \lambda_\mathbf{a} \mathcal{L}_{\mathbf{a}_\mathbf{c}} + \lambda_z \mathcal{L}_{\phi} + \lambda_{\text{rec}} \mathcal{L}^{\text{rec}}_\mathbf{c})
\) with their respective coefficients that control the relative importance of each loss term during optimization. We report the coefficients  \((10, 0.1, 0.1, 0.1, 0.0001, 10)\) according to our empirical studies. We use batch size \(bs=8\), and learning rate \(lr = 0.0001\) for training the PointNet-based backbones together with self-attention and reconstruction head. The residual flows are trained with \(lr_{flow} = 0.0003\). The feature embedding dimension is \(\mathbf{dim}(z_\mathbf{c}) = 240\). To ensure the normalizing flow learns from well-structured feature representations, we pre-train the encoder for 4000 iterations before incorporating the reconstruction and flow training.
\subsection{Simulation experiments}

\setlength{\tabcolsep}{3.5pt} 

\begin{table}[tbp]
\centering
\vspace{2mm}
\begin{tabular}{ccccccc}
\toprule

\textbf{Encoder} & \textbf{Loss} & 1- &  \textbf{AUSC} & \textbf{AUSE} & \textbf{AUSC}  & \textbf{AUSE}\\
&  &\textbf{cosine↓} &  \textbf{AL↓} & \textbf{AL↓} & \textbf{EP↓}  & \textbf{EP↓}\\
\midrule
& \(\mathcal{L}^{\text{cosine}}\) & 0.132 & / & / & / & / \\
{PointNet++} & \(\mathcal{L}^{\text{nll}}\) & 0.117 & 26.31 & 4.12 & / & / \\
& \(\mathcal{L}^{\text{Bl}}\) & 0.114 & 26.41 & 3.98 & 42.8 & 21.53 \\
\midrule

& \(\mathcal{L}^{\text{cosine}}\) & 0.127 & / & / & / & /  \\
{PointNet++}& \(\mathcal{L}^{\text{nll}}\) & 0.118 & 24.08 & 3.77 & / & /  \\
\&recon& \(\mathcal{L}^{\text{Bl}}\) & 0.105 & 23.63 & 3.51 & 41.88 & 21.77  \\
\midrule
\midrule
& \(\mathcal{L}^{\text{cosine}}\) & 0.125 & / & / & / & \\
{PointNeXt}& \(\mathcal{L}^{\text{nll}}\) & 0.116 & 24.82 & 4.07 & / & /  \\
& \(\mathcal{L}^{\text{Bl}}\) & 0.117 & 25.8 & 3.74 &35.61 & 13.3 \\
\midrule
& \(\mathcal{L}^{\text{cosine}}\) & 0.122 & / & / & / & /  \\
{PointNeXt}& \(\mathcal{L}^{\text{nll}}\) & 0.11 & 22.41 & \textbf{1.91}  & / & / \\
\&recon & \(\mathcal{L}^{\text{Bl}}\) & \textbf{0.083} & 22.23 &  1.93 & 32.54 &12.24 \\
\midrule
\midrule
& \(\mathcal{L}^{\text{cosine}}\) & 0.121 & / & / & / & /   \\
{Spotr}& \(\mathcal{L}^{\text{nll}}\) & 0.116 & 25.08 & 3.21 & / & / \\
& \(\mathcal{L}^{\text{Bl}}\) & 0.106 & 21.04 & 3.11 & 37.8 &  19.78 \\
\midrule

& \(\mathcal{L}^{\text{cosine}}\) & 0.11 & / & / & / & /\\
{Spotr}& \(\mathcal{L}^{\text{nll}}\) & 0.12 & 23.69 & 2.97 & / & /  \\
\&recon& \(\mathcal{L}^{\text{Bl}}\) & 0.097 & \textbf{21.02} & 3 & \textbf{28.45}  &  \textbf{10.52} \\
\bottomrule
\end{tabular}
 \caption{Baseline prediction results and uncertainty error. AL is aleatoric uncertainty and EP is epistemic uncertainty}
 \label{test}
\end{table} 

\paragraph{Evaluation metrics} 
Here, we focus on the prediction error and uncertainty estimation of baseline vectors. For baseline vector error, we applied the cosine similarity score as the first metric. Regarding uncertainty calibration error, we utilize the area under the sparsification curve (AUSC) \cite{ilg2018uncertainty} to evaluate the predicted uncertainty, along with the area under the sparsification error (AUSE), which measures the discrepancy between the AUSC and the actual error curve sorted by the defined error metric. Specifically, we sort the predicted contact points based on the uncertainty scores and calculate the error by top \(k\%\) samples with \(k\)-th percentile. As with the first metric, we take cosine similarity as the error metric for the actual error curve between the predicted contact baseline and its matched ground truth.

\begin{figure}
    \centering
    \vspace{2mm}
    \includegraphics[width=.9\linewidth]{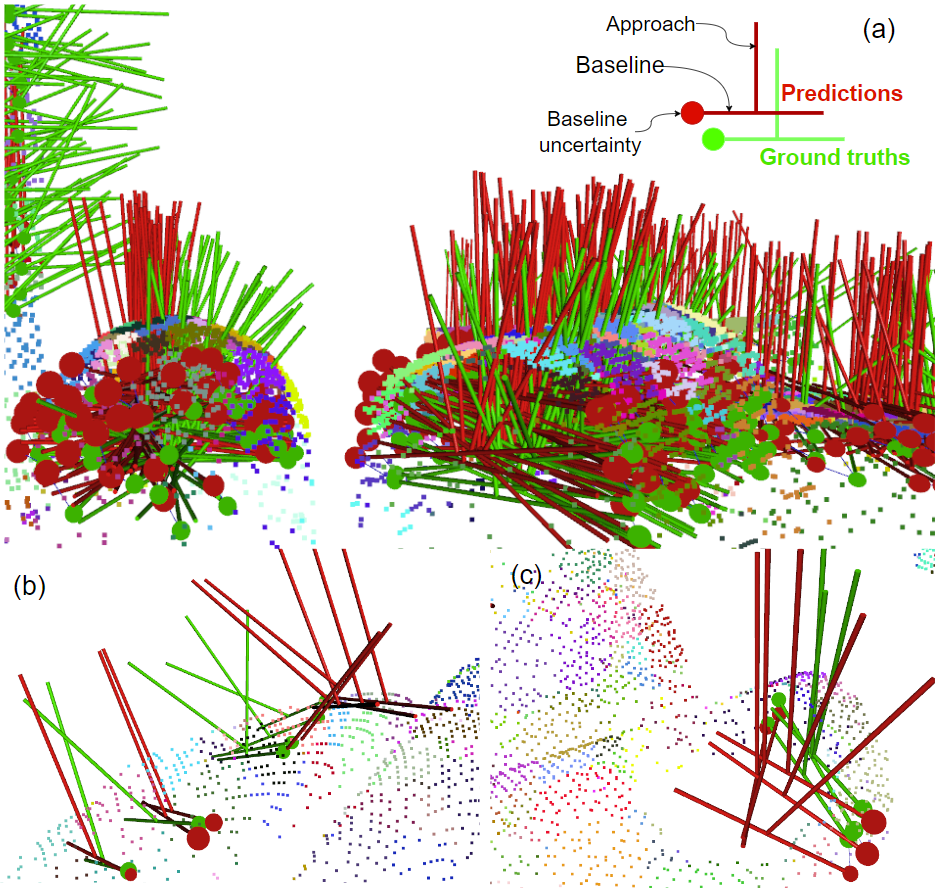}
    \caption{Visualization of predicted grasps (red) over the graspness threshold 0.5 compared with their nearest ground truths grasps (green). The smaller the sizes of red balls refer to higher total uncertainty.}
    \label{fig:vis}
\end{figure}
\paragraph{Results} We compared 3 recently proposed PointNet-based encoders with various architectures: PointNet++ \cite{ni2020pointnet++}, PointNeXt \cite{qian2022PointNeXt}, and Spotr \cite{park2023self} across pre-defined loss functions (\(\mathcal{L}^{\text{cosine}}\), \(\mathcal{L}^{\text{nll}}\), and \(\mathcal{L}^{\text{Bl}}\)).
Our results on the test data are recorded in Table \ref{test}. Across all encoders, Bayesian loss generally outperforms the other objectives across different encoders, yielding the lowest error, given the calibrated baselines by posterior update, which trains better aleatoric uncertainty prediction at the same time. Apart from that, the addition of the reconstruction task significantly boosts overall performance, particularly by enhancing feature informativeness. This is most notable in the PointNeXt, where the richer, more expressive features contribute to better uncertainty estimation and reliability. The complexity of PointNeXt allows it to leverage the reconstruction task to refine feature expressiveness at the same time, resulting in the highest accuracy of 0.083 and well-calibrated uncertainty predictions of 1.93. 

As qualitative results, we visualize the predicted grasps in Fig. \ref{fig:vis}. We observed that erroneous grasps, particularly those far from their matched ground truth, were typically associated with high predictive uncertainty (indicated by smaller red ball sizes), particularly in instances of geometric ambiguity caused by sparse point clouds and partial sensor observations (Fig.~\ref{fig:vis} (b, c)).

\subsection{Realworld experiments}
\paragraph{Experiment setups} In our real-world experiments, we aim to demonstrate the robustness of our method in handling uncertainty arising from novel scenarios. We utilize a UR10e robotic manipulator equipped with an Orbbec Femto Mega low-cost camera system and a Robotiq 2F85 gripper, as depicted in Fig.~\ref{fig:data}. In addition, we conduct two sets of experiments: one with in-distribution (ID) (cf. Fig.~\ref{fig:data} iii) objects, which are the same objects used for training data generation, and another one with out-of-distribution (OOD) items (cf. Fig.~\ref{fig:data} iv). The OOD items differ from the ID objects by various geometrical and physical properties such as a soft cube, a deformable glove and a cordless screwdriver. 
 
\begin{table}[t!]
\centering
\vspace{2mm}
\begin{threeparttable} 
\caption{Real-world experiment results}
\begin{tabular}{lcccc}
\toprule
\textbf{Condition} & \multicolumn{2}{c}{\textbf{Success Rate \%}} & \multicolumn{2}{c}{\textbf{Clearing Rate \%}} \\
\cmidrule(lr){2-3} \cmidrule(lr){4-5}
 & \textbf{ID} & \textbf{OOD} & \textbf{ID} & \textbf{OOD} \\
\midrule

\textbf{none}      & 39.2 & 45.5 & 61.1 & 66.7 \\
\textbf{recon}     & 60.6 & 56.1 & 83.3 & 77.8 \\
\textbf{recon + unc} & 72.7 & 65.0 & 88.9 & 88.9 \\

\bottomrule
\end{tabular} \label{table2}
\begin{tablenotes}
    \footnotesize
    \item * "unc" denote the uncertainty-based grasp sorting other than grasp score \(\text{P}(\mathbf{c})\)
\end{tablenotes}

\end{threeparttable}
\end{table}

\paragraph{Metrics and Results}
During the inference, we take the PointNext-based architecture as the baseline. The generated grasps are filtered using two strategies: i) the grasp quality \(\text{P}(\mathbf{c})\) or ii) total uncertainty \(\kappa'_0\). For execution, we randomly select the grasp from $10$-best scored grasps to avoid inference stagnation. We utilize grasp success rates and clearing rates as general performance metrics. in our real-world experiments. 
Three separate experiments were conducted using the same set of objects and scenes to enable quantitative comparison: 1) baseline with PointNeXt-based architecture, 2) with reconstruction, 3) with reconstructions and uncertainty. In total, we conducted 18 experiments. 

The experimental results are illustrated in Table.~\ref{table2}, which reveal significant improvements in both metrics when incorporating reconstruction and uncertainty-based grasp sampling. Specifically, the network trained by the reconstruction improves the geometric understanding of OOD objects, such as deformable or irregularly shaped items. Without these add-ons, the baseline model is sensitive to sensor noise and perception errors, leading to occasional misperception and suboptimal grasp performance even with ID objects. Our observations highlight the importance of both reconstruction and accurate uncertainty in improving grasp success and clearing rates, particularly in real-world scenarios.

\section{Conclusions}

We proposed vMF-Contact, an evidential grasp learning framework that leverages point-wise features to model both aleatoric and epistemic uncertainties with Bayesian treatment. We employed vMF prior to modeling directional uncertainty in the contact grasp representation and formulated a principle objective to implement posterior update. We further enhance feature expressiveness by applying patch-wise point reconstructions, which not only strengthens uncertainty quantification but also significantly enhances the model's generalization capability to handle OOD objects with varying scenes, viewpoints and sensor noises. Our method is proven to be valid across various PointNet-based architectures according to the evaluation. The real-world experiments further demonstrate the effectiveness of our methods incorporating reconstruction and uncertainty-based sorting, significantly improves grasp success. 

\bibliographystyle{IEEEtran}
\bibliography{bibfile}

\end{document}